\documentclass{llncs}

\usepackage{amssymb}
\usepackage{bm}
\usepackage{mathtools}
\usepackage{dsfont}
\usepackage{cite}
\usepackage{color,soul}

\newcommand{\R}[0]{\mathds{R}} 
\newcommand{\C}[0]{\mathds{C}} 

\providecommand{\mb}[1]{\mathbf{#1}}
\providecommand{\mbb}[1]{\boldsymbol{#1}}
\providecommand{\mbx}{\mb{x}}
\providecommand{\mby}{\mb{y}}

\providecommand{\pfpx}[2]{\frac{\partial{#1}}{\partial{#2}}}

\begin{document}

\title{A Deep Cascade of Convolutional Neural Networks for MR Image Reconstruction}
\titlerunning{CNN for MR Image Reconstruction}  
\author{Jo Schlemper\inst{1} \and Jose Caballero\inst{1} \and
Joseph V. Hajnal\inst{2} \and \\ Anthony Price\inst{2} \and Daniel Rueckert\inst{1}}

\authorrunning{Jo Schlemper et al.} 

\tocauthor{Jo Schlemper, Jose Caballero,
Joseph V. Hajnal, Anthony Price and Daniel Rueckert}

\institute{Imperial College London, London, United Kingdom\\
\email{jo.schlemper11@imperial.ac.uk}
\and King's College London, London, United Kingdom}

\maketitle              

\begin{abstract}
The acquisition of Magnetic Resonance Imaging (MRI) is inherently slow. Inspired
by recent advances in deep learning, we propose a framework for reconstructing
MR images from undersampled data using a deep cascade of convolutional neural
networks to accelerate the data acquisition process. We show that for Cartesian
undersampling of 2D cardiac MR images, the proposed method outperforms the
state-of-the-art compressed sensing approaches, such as dictionary
learning-based MRI (DLMRI) reconstruction, in terms of reconstruction error,
perceptual quality and reconstruction speed for both 3-fold and 6-fold
undersampling. Compared to DLMRI, the error produced by the method proposed is
approximately twice as small, allowing to preserve anatomical structures more
faithfully. Using our method, each image can be reconstructed in 23\,$ms$, which is fast enough to enable real-time
applications.

\keywords{Deep Learning, Convolutional Neural Network, Magnetic Resonance Imaging, Image Reconstruction}
\end{abstract}

\section{Introduction}

In many clinical scenarios, medical imaging is an indispensable diagnostic and research tool.
One such important modality is Magnetic Resonance Imaging (MRI), which is
non-invasive and offers excellent resolution with various contrast mechanisms to
reveal different properties of the underlying anatomy. However, MRI is associated
with a slow acquisition process. This is because data samples of an MR image are acquired sequentially in \emph{$k$-space} and the speed at which $k$-space can be traversed is limited by underlying MR physics. A long data acquisition procedures impose significant
demands on patients, making the tool expensive and less accessible. One possible
approach to accelerate the acquisition process is to undersample $k$-space, which in theory
provides an acceleration rate proportional to a reduction factor of a number of k-space traversals required.  However,
undersampling in $k$-space violates the Nyquist-Shannon theorem and generates
aliasing artefacts when the image is reconstructed. The main challenge in this case is to find an algorithm that takes into account the undersampling undergone and can compensate missing data with a-priori knowledge on the image to be reconstructed.

Using Compressed Sensing (CS), images can be reconstructed from sub-Nyquist
sampling, assuming the following: firstly, the acquired images must be \emph{compressible}, i.e.
they have a sparse representation in some transform domain. Secondly, one must ensure \emph{incoherence} between the sampling and sparsity domains to guarantee that the reconstruction problem has a unique solution and that this solution is attainable. In practice, this can be achieved with random sub-sampling of $k$-space, which translates aliasing patterns in the image domain into patterns that can be regarded as correlated noise. Under such assumptions, images can then be
reconstructed through nonlinear optimization or iterative algorithms. The class of methods which applies CS to the MR reconstruction problem is termed CS-MRI \cite{Lustig2008}. A natural extension of these has been to enable more flexible representations with \emph{adaptive} sparse modelling, where one attempts to
obtain the optimal representation from data directly. This can be done by
exploiting, for example, dictionary learning (DL) \cite{Ravishankar2011}. 

To achieve more aggressive undersampling, several strategies can be considered. One way is to further exploit the inherent redundancy of the MR data. For example, in dynamic imaging, one can make use of spatio-temporal redundancies \cite{Caballero2014}, \cite{Jung2007}, \cite{Quan2016}. Similarly, when imaging a full 3D volume, one exploit redundancy from adjacent slices \cite{Hirabayashi2015}. An alternative approach is to exploit sources of explicit redundancy of the data and solve an overdetermined system. This is the fundamental assumption underlying parallel imaging \cite{Uecker2014}. Similarly, one can make use of multi-contrast information 
\cite{Huang2014a} or the redundancy generated by multiple filter responses of the image \cite{XiPeng2015}. These explicit redundancies can also be used to complement the sparse modelling of inherent redundancies \cite{Jin2015}, \cite{Liang2009a}.

Recently, deep learning has been successful at tackling many computer vision problems. Deep neural network architectures, in particular convolutional
neural network (CNN), are becoming the state-of-the-art technique for various
imaging problems including image classification \cite{He2015}, object localisation \cite{Ren2015} and image
segmentation \cite{Ronneberger2015}.  Deep architectures are capable of extracting features from data to build increasingly abstract representations that are useful for the end-goal being considered, replacing the traditional approach of carefully hand-crafting features and algorithms. For example, it has already been demonstrated that CNNs outperform sparsity-based methods in super-resolution \cite{Dong2016}, not only for its quality but also in terms of the reconstruction speed \cite{Shi_2016_CVPR}. One of the contributions of our work is to explore the application of CNNs in undersampled MR reconstruction and investigate whether they can exploit data redundancy through learned representations. In fact, CNNs have already been applied to compressed sensing from random Gaussian measurements
\cite{Kulkarni2016a}. Despite the popularity of CNNs, there has only been preliminary research on CNN-based MR image reconstruction \cite{NIPS2016_6406}, \cite{wang2016b}, hence the applicability of CNNs to this problem is yet to be qualitatively and quantitatively assessed in detail. 

In this work we consider reconstructing 2D static images with Cartesian
sampling using CNNs. Similar to the formulations in CS-MRI, we view the
reconstruction problem as a de-aliasing problem in the image domain. However,
reconstructing an undersampled MR image is challenging because the images
typically have low signal-to-noise ratio, yet often high-quality
reconstructions are needed for clinical applications. To resolve this issue, we propose a very deep network architecture which forms a cascade of
CNNs. Our cascade network closely simulates the iterative reconstruction of
DL-based methods, however, our approach allows end-to-end optimisation of the reconstruction algorithm. We show that under the Cartesian undersampling scheme, our CNN approach is capable of producing high-quality reconstructions of 2D cardiac MR images, outperforming DL-based MRI reconstruction  (\emph{DLMRI}) \cite{Ravishankar2011}. Moreover, using the proposed method, each images can be reconstructed in about 23\,$ms$, which enables the real-time applications.

\section{Problem Formulation}

Let $\mbx \in \C^N$ represent a complex-valued MR image composed of
$\sqrt{N}\times \sqrt{N}$ pixels stacked as a column vector. Our problem is to
reconstruct $\mbx$ from $\mby \in \C^M$, the measurements in $k$-space, such
that:

\begin{equation}
  \mby = \mb{F}_u \mbx
  \label{eq:cs_basic}
\end{equation}

Here $ \mb{F}_u \in \C^{M\times N}$ is an undersampled Fourier encoding matrix.
For undersampled $k$-space measurements ($M << N$), the
system of equations (\ref{eq:cs_basic}) is underdetermined and hence the
inversion process is ill-defined. In order to reconstruct $\mbx$, one must exploit a-priori knowledge of its properties, which can be done by formulating an unconstrained optimisation problem:

\begin{equation}
  \label{eq:sparse_coding}
\begin{aligned}
& \underset{\mbx}{\text{min.}}
& & \mathcal{R}(\mbx) + \lambda \| \mby - \mb{F}_u \mbx \|^2_2
\end{aligned}
\end{equation}

$\mathcal{R}$ expresses regularisation terms on $\mbx$ and $\lambda$ allows the adjustment of data fidelity based on the noise level of the acquired
measurements $\mby$. For CS-based methods, the regularisation terms $\mathcal{R}$ typically involve $\ell_0$ or
$\ell_1$ norms in the sparsifying domain of $\mbx$. Our formulation is inspired by DL-based reconstruction approaches, in which the problem is formulated as:

\begin{equation}
  \label{eq:dl}
\begin{aligned}
& \underset{\mbx, \mb{D}, \{ \mbb{\gamma}_i\} }{\text{min.}}
& & \sum_i \left(\|\mb{R}_i\mb{x} - \mb{D} \mbb{\gamma}_i \|_2^2 + \nu \|\mbb{\gamma}_i \|_0 \right) + \lambda \| \mby - \mb{F}_u \mbx \|^2_2
\end{aligned}
\end{equation}


Here $\mb{R}_i$ is an operator which extracts an image patch at $i$, $\mbb{\gamma}_i$ is the corresponding sparse code with respect to a dictionary $\mb{D}$. In this approach, the regularisation terms enforce $\mbx$ to be approximated by the reconstructions from the sparse code of patches. By taking the same approach, for our CNN formulation, we enforce $\mb{x}$ to be well-approximated by the CNN reconstruction:

\begin{equation}
  \begin{aligned}
    \label{eq:cnn_rec}
    & \underset{\mbx, \boldsymbol{\theta}}{\text{min.}}
    & & \| \mbx - f_{\text{cnn}}(\mbx_u | \boldsymbol{\theta}) \|^2_2 + \lambda \| \mb{F}_u \mbx - \mby \|_2^2 \\ 
  \end{aligned}
\end{equation}

Here $f_{\text{cnn}}$ is the forward mapping of the CNN parameterised by $\boldsymbol{\theta}$, which takes in the zero-filled reconstruction $\mbx_u = \mb{F}^{H}_u \mby$ and directly produces a reconstruction as an output. Since $\mb{x}_u$ is heavily affected by aliasing from sub-Nyquist sampling, the CNN reconstruction can therefore be seen as solving de-aliasing problem in the image domain. 

The approach of eq. (\ref{eq:cnn_rec}), however, is limited in the sense that the CNN reconstruction and the data fidelity are two independent terms. In particular, since the CNN operates in the image domain, it is trained to
reconstruct the image without a-priori information of the acquired data in $k$-space. However,
if we already know some of the $k$-space values, then the CNN should be discouraged from modifying them. Therefore, by incorporating the data fidelity in the learning stage, the CNN should be able to achieve better reconstruction. This means that the output of the CNN is now conditioned on $\Omega$, an index set indicating which $k$-space measurements have been sampled in $\mby$. Then, our final reconstruction is given simply by the output, $\mbx_\text{cnn} = f_{\text{cnn}}(\mbx_u | \boldsymbol{\theta}, \lambda, \Omega)$. Given training data
$\mathcal{D}$ of input-target pairs $(\mbx_u, \mbx_t)$, we can train the CNN to produce an output that attempts to accurately reconstruct the fully-sampled data by minimising an objective function:

\begin{equation}
\mathcal{L}(\boldsymbol{\theta}) = \sum_{(\mbx_u, \mbx_t) \in \mathcal{D}} \ell \left( \mbx_t,
\mbx_\text{cnn} \right)\label{eq:cnn_objective}
\end{equation}

where $\ell$ is a loss function. In this work, we consider an element-wise squared loss, which is given by $\ell \left(\mbx_t, \mbx_\text{cnn} \right) = \| \mbx_t - \mbx_\text{cnn} \|_2^2$.


\section{Data Consistency Layer}

In order to incorporate the data fidelity in the network architecture, we first note the following: for a fixed $\boldsymbol{\theta}$, eq. (\ref{eq:cnn_rec}) has a closed-form solution in $k$-space, given as in \cite{Ravishankar2011}:

\begin{equation}
\label{eq:dc_step_in_k}
\hat{\mbx}_{\text{rec}}(k) =
    \begin{cases}
      \hat{\mbx}_{\text{cnn}}(k) & \text{if } k \not \in \Omega\\
      \frac{\hat{\mbx}_{\text{cnn}}(k) + \lambda \hat{\mbx}_u(k)}{1+\lambda} & \text{if } k \in \Omega
    \end{cases}
\end{equation}

where $\hat{\mbx}_{\text{cnn}} = \mb{F} f_{\text{cnn}}(\mbx_u | \boldsymbol{\theta})$, $\hat{\mbx}_u = \mb{F}\mbx_u$ and $\mb{F}$ is the Fourier encoding matrix. The final image is reconstructed by applying the inverse of the encoding matrix $\mbx_{\text{rec}} = \mb{F}^{-1}
\hat{\mbx}_{\text{rec}}$. In the noiseless setting (i.e.  $\lambda \to \infty$), we simply replace the $i$th predicted coefficient by the original coefficient if it
has been sampled. For this reason, this operation is called \emph{data consistency step} in $k$-space (DC). 

Since the DC step has a simple expression, we can in fact treat it as a layer operation of the network, which we denote as \emph{DC layer}. When defining a layer of a network, the rules for forward and backward passes must be specified. This is because CNN training can effectively be performed through stochastic gradient descent, where one updates the network parameters $\boldsymbol{\theta}$ to minimise the objective function $\mathcal{L}$ by descending along the direction given by the derivative $\partial \mathcal{L} / \partial \boldsymbol{\theta}^T$. 
 For this, it is necessary to define the gradients of each network layer relative to the network's output. In practice, one uses an efficient algorithm called \emph{backpropagation} \cite{backprop}, where the final gradient is given by the product of all the Jacobians of the layers contributing to the output. Hence, in general, it suffices to specify a layer operation $f_L$ for the forward pass and derive the Jacobian of the layer with respect to the layer input $\partial f_{L}
 / \partial \boldsymbol{\mbx}^T$ 
 for the backward pass. 

\paragraph{Forward pass} The data consistency in $k$-space can be simply decomposed into three operations: Fourier transform, data consistency and inverse Fourier transform. In our case, we take our Fourier transform to be a two-dimensional (2D) discrete Fourier transform (DFT) of the 2D image representation of $\mbx$, which is written as $\hat{\mbx} = \mb{F} \mbx$ in matrix form. The inverse transformation is defined analogously, where $\mbx = \mb{F}^{-1} \hat{\mbx}$. The data consistency $f_{dc}$ performs the element-wise operation defined in eq. (\ref{eq:dc_step_in_k}). We can write it in matrix form as:

\begin{equation}
f_{dc}(\hat{\mbx}, \hat{\mbx}_u, \lambda) = \boldsymbol{\Lambda}\hat{\mbx} + \frac{\lambda}{1 + \lambda} \hat{\mbx}_u\label{eq:fill_mat}
\end{equation}

Here $\bm{\Lambda}$ is a diagonal matrix of the form:

\begin{equation}
\bm{\Lambda}_{kk} =
\begin{cases}
  1 & \text{if } k \not \in \Omega \\
  \frac{1}{1+\lambda} & \text{if } k \in \Omega
\end{cases}\label{eq:fill_matrix}
\end{equation}

Combining the three operations defined above, we can obtain the forward pass of the layer performing data consistency in $k$-space: 

\begin{equation}
f_{L}(\mbx, \hat{\mbx}_u, \lambda) = \mb{F}^{-1}\mb{\Lambda}\mb{F}\mbx + \frac{\lambda}{1 + \lambda} \mb{F}^{-1}\hat{\mbx}_u\label{eq:dc_fnc}
\end{equation}

\paragraph{Backward pass} In general, one requires \emph{Wirtinger calculus} to derive a gradient in complex domain\cite{FaijulAmin2012}, however, in our case, the derivation greatly simplifies due to the linearity of the DFT matrix and the data consistency operation. The Jacobian of the DC layer with respect to the layer input $\mbx$ is therefore given by:
\begin{equation}
\pfpx{{f_{L}}}{\mbx^T} = \mb{F}^{-1}\mb{\Lambda}\mb{F}
\end{equation}
There are several points that deserve further explanation: firstly, unlike many other applications where CNNs process real-valued data, MR images are complex-valued and the network needs to account for this. One possibility would be to design the network to perform complex-valued operations. A simpler approach, however, is to accommodate the complex nature of the data with real-valued operation in a dimensional space twice as large (i.e. we replace $\C^N$ by $\R^{2N}$).  In the latter case, the derivations above still hold due to the fundamental
assumption in Wirtinger calculus. Secondly, even though the DC layer does not have any additional parameters to be optimised, it allows end-to-end training of CNN, hence benefiting our final reconstruction. 

\section{Cascading Network}

For CS-based methods, in particular for DLMRI, the optimisation problem is
solved using a coordinate-descent type algorithm, alternating between the de-aliasing
step and the data consistency step until convergence. In contrast, with CNNs, we are performing one step de-aliasing and the same network cannot be used to de-alias iteratively.
While CNNs may be powerful enough to learn one step reconstruction,
such network could indicate signs of overfitting, unless we have vast
amounts of training data. In addition, training such networks may require a long
time as well as careful fine-tuning steps. It is therefore best to be able to
use CNNs for iterative reconstruction approaches.

A simple solution is to train a second CNN which learns to reconstruct from
the output of the first CNN. In fact, we can concatenate a new CNN on
the output of the previous CNN to build extremely deep networks which iterate
between intermediate de-aliasing and the data consistency reconstruction. We
term this a \emph{cascading network}. In fact, one can essentially
view this as unfolding the optimisation process of DLMRI. If each CNN expresses the dictionary learning reconstruction step, then the cascading CNN
can be seen as a direct extension of DLMRI, where the whole reconstruction
pipeline can be optimised from training. 

\section{Architecture and Implementation}

\begin{figure}[!t]
  \centering
  \includegraphics[width=1\textwidth]{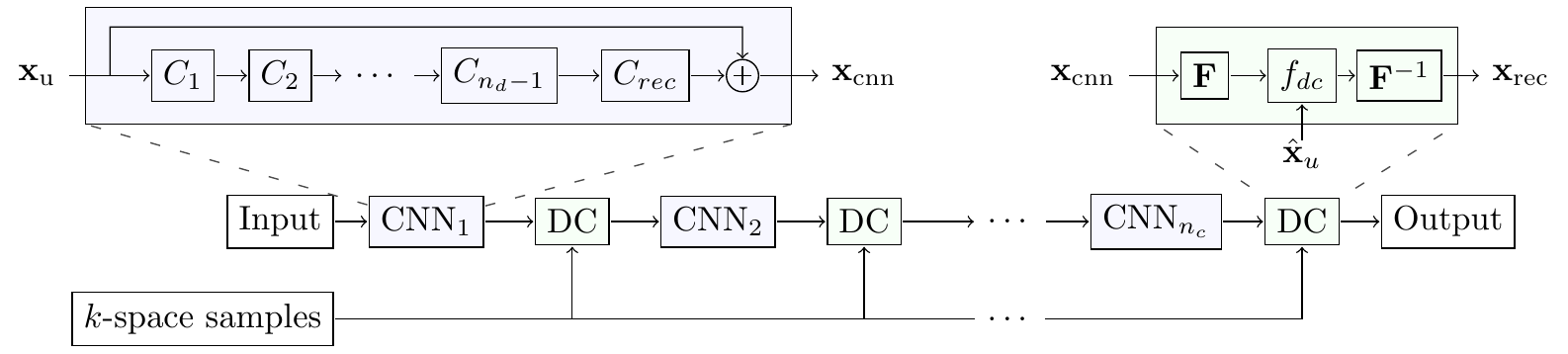}
  \caption{A cascade of CNNs. The depth of architecture and the depth of cascade is denoted by $n_d$ and $n_c$ respectively.}
\label{fig:custom_resnet_nd_nc} 
\end{figure}

Incorporating all the new elements mentioned above, we can devise our cascading network
architecture. Our CNN takes in a two-channeled image $\R^{\sqrt{n}\times \sqrt{n}\times2}$, where each channel stores real and imaginary parts of the undersampled image. Based on literature, we used the following network architecture for CNN, illustrated in Figure \ref{fig:custom_resnet_nd_nc}: it has $n_d-1$ convolution layers $C_i$, which are all followed by Rectifier Linear Units (ReLU) as a choice of nonlinearity. For each of them, we used a kernel size $k=3$ \cite{Szegedy2015} and the number of filters were set to $n_f = 64$. The network is followed by another convolution layer $C_{\text{rec}}$ with kernel size $k = 3$ and $n_f = 2$,  which projects the extracted representation back to image domain. We also used \emph{residual connection} \cite{He2015a}, which sums the output of the CNN module with its input. Finally, we form a cascading network by using the DC layers interleaved with the CNN reconstruction modules. For our experiment, we chose $n_d=5$ and $n_c=5$. We found that our choice of hyperparameters work sufficiently well, however, by no means were they optimised. Hence the result is likely to be improved by changing the architecture and varying the parameters such as kernel size and stride \cite{Ronneberger2015}, \cite{Yu2016}. 

As mentioned, pixel-wise squared error was used as our objective function. The
minibatch size was set to 10, however, for the deeper models with large number
of cascades, the minibatch size was reduced to fit the model on a single GPU memory.
We initialised our network weights using He initialisation \cite{He2015a}. Adam
optimiser \cite{Kingma2014} was used to train all models, with the parameters
$\alpha = 10^{-4}, \beta_1 = 0.9$ and $\beta_2 = 0.999$.
We also added $\ell_2$ weight decay of $10^{-7}$. The network was implemented in Python using Theano and Lasagne libraries.

\section{Experimental Results}
\subsection{Setup}

\paragraph{Dataset}

Our method was evaluated using the cardiac MR dataset used in \cite{Caballero2014}, consisting of 10 fully sampled short-axis cardiac cine MR scans. Each scan contains a single slice SSFP acquisition with 30 temporal frames with a $320 \times 320$ mm field of view and 10 mm slice thickness. The raw data consists of 32-channel data with sampling matrix size $192\times 190$, which was zero-filled to the matrix size $256\times256$. The data was combined into a single complex-valued image using SENSE \cite{Pruessmann1999} with no undersampling, retrospective gating and the coil sensitivity maps normalised to a body coil image. The images were then retrospectively undersampled using Cartesian undersampling
masks, where we fully sampled along the frequency-encoding direction but undersample in the phase-encoding direction. The strategy was adopted from \cite{Jung2007}: for each frame we acquired eight lowest spatial frequencies. The sampling probability of other frequencies along the phase-encoding direction was determined by a zero-mean Gaussian distribution. The acceleration rates are stated with respect to the matrix size of the raw data. 
Note that similarly to previous studies, \cite{Ravishankar2011}, \cite{Caballero2014}, since the raw data was combined prior to the simulation, the coil sensitivities were not directly addressed in our reconstruction. This is set for future investigation, where we plan to incorporate the explicit redundancy created by parallel imaging into our model.

Although the dataset is a dynamic sequence, we restrict our experiments to the
2D case only. Therefore, each time frame was treated as an independent image,
yielding a total of 300 images. 
We found that applying rigid transformations as a data augmentation was crucial, as without it, the network quickly overfitted the training data. 
Moreover, for a fixed undersampling rate, we generated an undersampling mask on-the-fly to allow the network to learn diverse patterns of aliasing artefact.

\paragraph{Metric}

We evaluated our method by reconstructing undersampled images from 3-fold and 6-fold acceleration rates. We used mean squared error (MSE) as our quantitative measure. During our experiment, we noticed that even for the same undersampling rate, different undersampling masks yield considerable differences in the reconstruction's signal-to-noise. To take this into consideration for fair comparison, we assigned an arbitrary but fixed undersampling mask for each image in test data. Apart from the quantitative measure, we also inspected the visual aspect of the reconstructed images for qualitative assessment.

\paragraph{Models} 

For CNN, we selected the hyperparameters described above. To ensure a fair comparison, we
reported the aggregated test result from 2-fold cross-validation (i.e. train on five subjects and test on the other five). For each iteration of the cross validation, the network was initialised using He initialisation, trained end-to-end. For 6-fold undersampling, we initialised the network using the parameters obtained from the trained models from 3-fold acceleration and fine-tuned using Adam optimiser. Each network converged within 3 days on GeForce TITAN X.

We compared our method to DLMRI, a representative of the state-of-the-art CS-based methods. 
 For DLMRI, we simplified the implementation of DLTG from \cite{Caballero2014}, with patch size $6 \times 6$. We switched off any de-aliasing along the temporal axis. Since DLMRI
is quite time consuming, in order to obtain the results within a reasonable amount of
time, we trained a joint dictionary for all time frames within the subject and reconstructed them
in parallel. Note that we did not observe any decrease in performance from
this approach. For each subject, we ran 400 iterations and obtained the final
reconstruction.

\subsection{Results}
\label{sec:vs_dl_2d}

\begin{table}[!t]
\renewcommand{\arraystretch}{1.3}
\caption{DLMRI vs. CNN across 10 scans}
\label{table:vs_dl}
\centering
\begin{tabular}{c|c||c|}
\cline{2-3}
  & \multicolumn{1}{c||}{3-fold} & \multicolumn{1}{c|}{6-fold} \\
\hline
\multicolumn{1}{|c|}{Models} & MSE (SD) $\times 10^{-3}$ & MSE (SD) $\times 10^{-3}$ \\
\hline
\multicolumn{1}{|c|}{DLMRI} & 2.12 (1.27) & 6.31 (2.95) \\
\hline
\multicolumn{1}{|c|}{CNN} & \textbf{0.89} (\textbf{0.46}) & \textbf{3.42} (\textbf{1.65}) \\
\hline
\end{tabular}
\end{table}

\begin{figure}[!t]
  \centering
  \includegraphics[width=\textwidth]{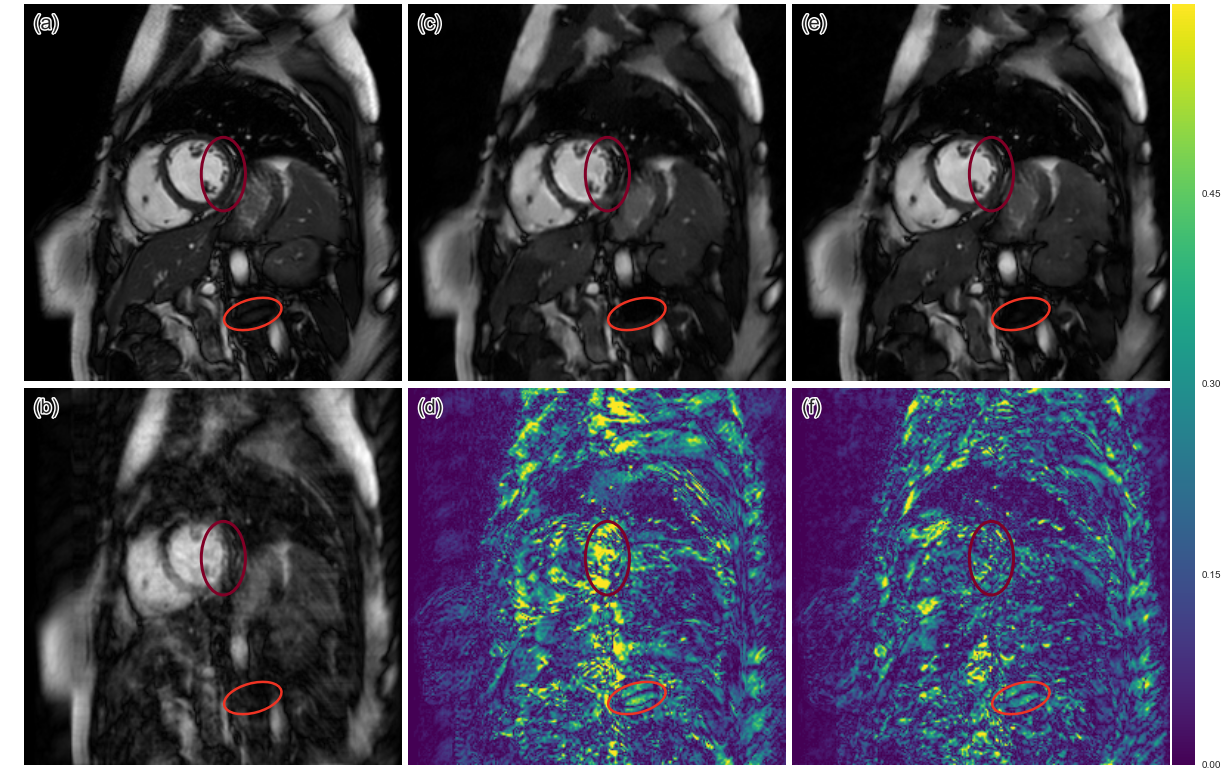}
  \caption{The comparison of reconstruction from DLMRI and CNN. (a) The original,
    (b) 3x undersampled, (c)-(d) DLMRI reconstruction and its error map $\times
    5$ and (e)-(f) CNN reconstruction and its error map $\times 5$. }
  \label{fig:tmi2_ac4_detailed}
\end{figure}

\begin{figure}[!t]
  \centering
  \includegraphics[width=\textwidth]{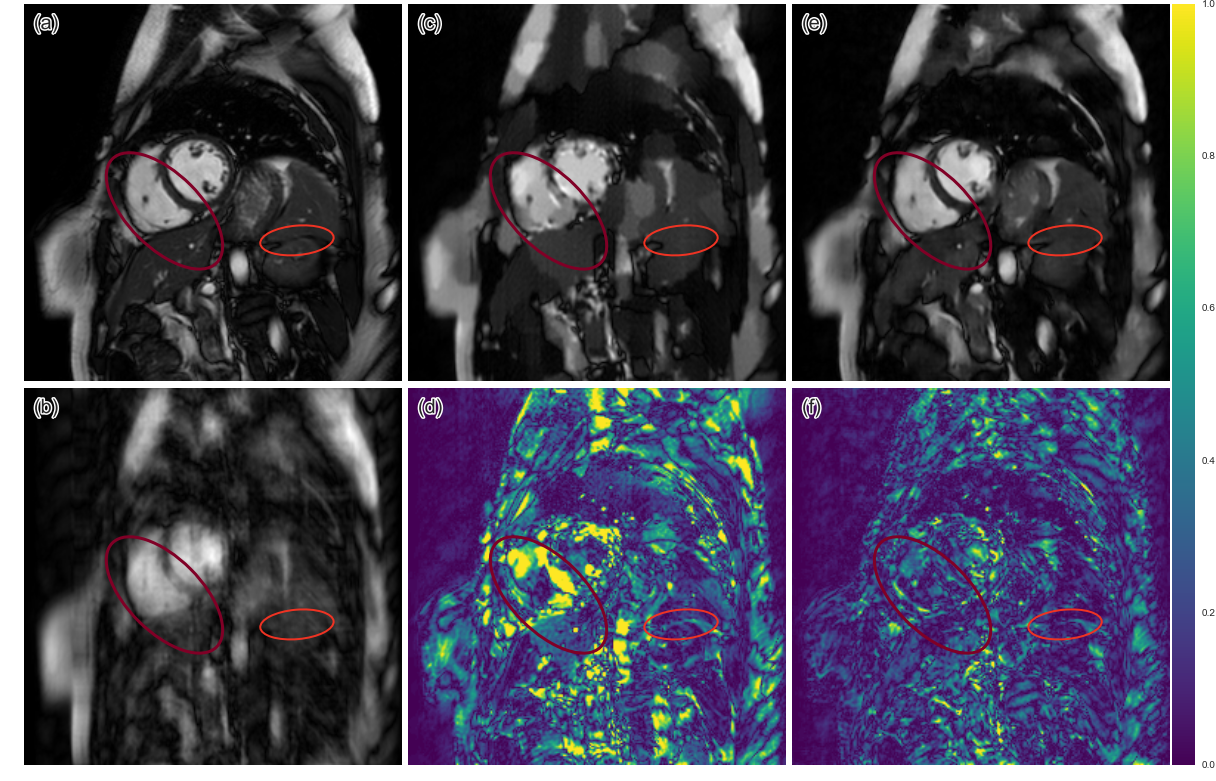}
  \caption{The comparison of reconstructions from DLMRI and CNN. (a) The original,
    (b) 6x undersampled, (c)-(d) DLMRI reconstruction and its error map $\times 5$ and
    (e)-(f) CNN reconstruction and its error map $\times 5$.}
\label{fig:tmi2_ac8_detailed}
\end{figure}

The means of the reconstruction errors across 10 subjects are summarised in table \ref{table:vs_dl}. For both 3-fold and 6-fold acceleration, one can see that CNN consistently outperformed
DLMRI, and that the standard deviation of the
error made by CNN was smaller. 
The reconstruction from 3-fold acceleration can be found in Figure \ref{fig:tmi2_ac4_detailed}. It can be seen that the CNN approach produced a smaller overall error. The CNN reconstruction produced a more homogeneous
reconstruction. On the other hand, DLMRI gave a blocky
reconstruction. In some cases, both CNN and DLMRI suffered
from small losses of important anatomical structures in their reconstructions (orange), but CNN was able to recover more details (red). The reconstructions from 6-fold acceleration is in Figure \ref{fig:tmi2_ac8_detailed}. Although both
methods suffered from significant loss of structures (orange), CNN was still capable of better preserving the texture than DLMRI (red). On the other hand, DLMRI created
extremely block-like artefacts due to over-smoothing. 6x undersampling for these images typically approaches the limit of sparsity-based methods, however, CNN was able to predict some anatomical details which was not possible by DLMRI. This could be due to the fact that CNN has more free parameters to tune with, allowing the network to learn complex but more accurate transformations of data.  

\paragraph{Comparison of Reconstruction Speed} While training CNN is time consuming, once it is trained, the inference
can be done extremely quickly on a GPU. Reconstructing each slice took  $23 \pm 0.1$
milliseconds on GeForce GTX 1080, which enables real-time applications. To produce the above results, DLMRI took
about $6.1 \pm 1.3$ hours per subject on CPU. Even though we do not have a GPU implementation of DLMRI, it is expected to take longer than 23ms because DLMRI requires dozens of iterations of dictionary learning and sparse coding steps. Using a fixed, pre-trained dictionary could remove this bottleneck, in exchange of lowering the reconstruction capacity. 

\section{Discussion and Conclusion}

In this work, we evaluated the applicability of CNNs for the MR image
reconstruction problem. From the experiment, we have
shown that using the network with interleaved data consistency stages, we can obtain a model which can reconstruct images sufficiently well. The CS framework offers mathematical guarantee for the
signal recovery, which makes the approach appealing in theory as well as  in practice even though the required sparsity cannot be genuinely achieved in medical imaging. However, even though this is not the case for CNNs, we have empirically shown that a CNN-based approach can outperform DL-based MR reconstruction.  In addition, at very aggressive undersampling rates, the CNN method was capable of reconstructing most of the anatomical structures more accurately, while CS-based methods do not guarantee such behaviour. 

The limitation of this work is that the data was first reconstructed by SENSE, which was then used to simulated the acquisition process. It is, however, more practical to consider images with sensitivity map of the surface coils, which allows the model to be used for parallel imaging reconstruction directly. In fact, a better approach is to exploit the redundancy of the coil sensitivity maps and combine directly into our model, which will be addressed in our future work. 

In this work, we were able to show that the network can be trained using arbitrary Cartesian undersampling masks of the fixed sampling rate rather than selecting a fixed number of undersampling masks for training and testing. This suggests that the network was capable of learning a generic strategy to de-alias the images. A further investigation should consider how tolerant the network is for different undersampling rates. Furthermore, it is interesting to consider other sampling patterns such as radial and spiral trajectories. As these trajectories provide different properties of aliasing artefacts, a further validation is appropriate to determine the flexibility of our approach.

Finally, Although CNNs can only learn local representations which should not affect global structure, it remains to be determined how the CNN approach operates when there is a pathology present in images, or other more variable content. We have performed a two-fold cross-validation to ensure that the network can handle unseen data acquired through the same acquisition protocol. Generalisation properties must be evaluated carefully on a larger dataset, however, CNNs are flexible in a way such that one can incorporate application specific priors to its objective to allocate more importance on preserving any features of interest in the reconstruction, provided that such expert knowledge is available at training time. For example, analysis of cardiac images in clinical settings often employs segmentation and/or registration. Multi-task learning is a promising approach to further improve the utility of CNN-based MR reconstructions.

\section*{Acknowledgment}

The work was partially funded by EPSRC Programme Grant (EP/P001009/1).

%
%
\bibliographystyle{splncs03}
\bibliography{library.bib}







\end{document}